\documentclass{svproc}
\usepackage{cite}
\usepackage[hyphens]{url}
\usepackage{graphicx}
\begin{document}
\mainmatter              
\title{Causal-BERT : Language models for causality detection between events expressed in text}

\titlerunning{Causal-BERT}  
%

\author{Vivek Khetan\inst{1}, Roshni Ramnani\inst{1}, Mayuresh Anand\inst{2}\thanks{This work was performed during an internship at Accenture Labs, SF.}, Subhashis Sengupta\inst{1}, Andrew E. Fano\inst{1}}
\institute{Accenture Labs  \and
  UC Santa Barbara \\
  \{vivek.a.khetan, roshni.r.ramnani , shubhashis.sengupta, andrew.e.fano,\}@accenture.com \\ mayuresh@ucsb.edu}

\authorrunning{Khetan et al.} 


\maketitle              


\begin{abstract}
Causality understanding between events is a critical natural language processing task that is helpful in many areas, including health care, business risk management, and finance. On close examination, one can find a huge amount of textual content both in the form of formal documents or in content arising from social media like Twitter, dedicated to communicating and exploring various types of causality in the real world. Recognizing these "Cause-Effect" relationships between natural language events continues to remain a challenge simply because it is often expressed implicitly. Implicit causality is hard to detect through most of the techniques employed in literature and can also, at times be perceived as ambiguous or vague. Also, although well-known datasets do exist for this problem, the examples in them are limited in the range and complexity of the causal relationships they depict especially when related to implicit relationships. Most of the contemporary methods are either based on lexico-semantic pattern matching or are feature-driven supervised methods. Therefore, as expected these methods are more geared towards handling explicit causal relationships leading to limited coverage for implicit relationships, and are hard to generalize. In this paper, we investigate the language model's capabilities for causal association among events expressed in natural language text using sentence context combined with event information, and by leveraging masked event context with in-domain and out-of-domain data distribution. Our proposed methods achieve the state-of-art performance in three different data distributions and can be leveraged for extraction of a causal diagram and/or building a chain of events from unstructured text. 
\keywords{Causal Relations, Cause-Effect, Causality Extraction, Language Models, Causality in natural language text}
\end{abstract}
\section{Introduction}
Recent advances in machine learning have enabled progress on a range of tasks. Causal and common sense reasoning, however, has not benefited proportionately. Causal reasoning efforts can be grouped into the task of causal discovery \cite{Pearl2009} and the task of causality understanding for events described in natural language text \cite{Hassanzadeh2019}. Causality detection in natural language is often simplified/described as detection of "Cause-Effect" relation between two events, where an event is expressed as a nominal, phrase, or short span of text in the same or different sentences \cite{Luo2016}. 


Understanding causal association between everyday events is very important for common sense language understanding (e.g., "John was late; Bob got angry.") as well as causal discovery. For our purposes, however, we will focus on commercial industry applications (e.g., “The phone was dropped in a lake; hence the warranty is void.”). Understanding the potential "Cause-Effect" relationship between events can form the basis of several tasks such as binary causal question-answering (\cite{Hassanzadeh2019} and \cite{Sharp2016}), a plausible understating of the relationship between everyday activities (\cite{Gordon2011} and \cite{Luo2016}), adverse effects of drugs, and decision-support tasks \cite{Hassanzadeh2020}.  
\begin{figure}[hbt!]
    \centering
    \includegraphics[width=9cm]{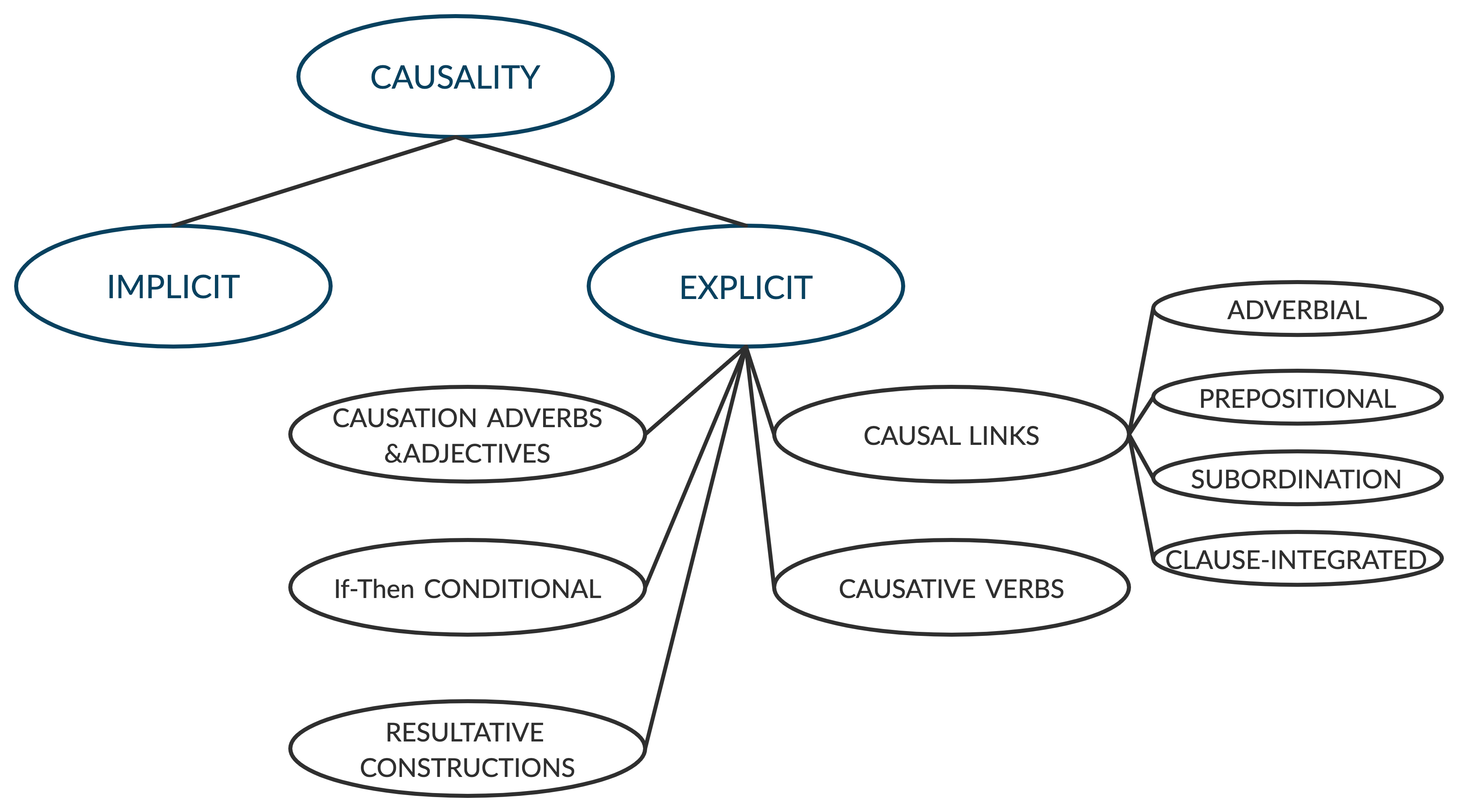}
    \caption{Causality in natural language}
    \label{fig:causality-in-text}
\end{figure}

Most of the current methods for causality understanding write linguistic pattern-matching rules or use careful feature engineering to train a supervised machine learning algorithm \cite{Girju2002}. They often lack coverage and are hard to scale for an unseen sequence of events. Identifying the relationship often depends on the whole sentence structure, along with its semantic features. Additionally, it is challenging to decompose every event and the rest of the expression completely in their lexical representation \cite{Chang1991}. Furthermore, due to the limited availability of comprehensive annotated datasets for "Cause-Effect," only a few deep learning-based approaches have been proposed \cite{Dasgupta2019}. 

This work focuses on understanding the causality between events expressed in natural language text. The intent is simply to identify possible causal relationships between marked events implied by a given sequence of text. It is not to determine the validity of those relationships nor to provide a highly nuanced representation of those relationships. The events can be a nominal, a phrase, or a span of text in the given event statement. We built network-architectures on top of language models that leverage overall sentence context, event' context ,and events' masked context for causality detection between expressed events. Pre-training on out of domain data-distribution helped our model gain performance, showing that language models can learn implicit structural representation in natural language text that is in line with the findings of \textit{Jawahar, Goldberg and Gururangan} [\cite{Jawahar2020}, \cite{Goldberg2019} and \cite{Gururangan2020}]. We curated the datasets from Semeval 2010 task 8 \cite{Hendrickx2010}, Semeval 2007 task 4 \cite{Beamer2007}\footnote{We use terms Semeval 2010 and Semeval 2007 for  Semeval 2010 Task 8 and Semeval 2007 Task 4 respectively}, ADE \cite{Gurulingappa2012} corpus. All the sentences in the three mentioned datasets were annotated with the provided event description and were given "Cause-Effect" or "Other" labels for each pair of event interactions. \\
\noindent Our contributions can be summarised as follows: 
\begin{itemize}
\item We curated three different datasets from publicly available tasks for causality detection in natural language text 
\item We showed that language models can be trained to learn causal interaction between events expressed in natural language text
\item We showed that that pre-training of language models using out-of-domain text expression improves the performance in low resource causality detection
\end{itemize}

\section{Event-structure and causality in natural language text}
Interaction between events expressed in natural language text is not just dependent on how implicit or explicit the events are mentioned in the expression, but also on the sentence's arguments and the lexical representation of those arguments \cite{Chang1991}.
Predicting causal interaction between events is a complex task because it depends on interactions between the particular linguistic expression of information, semantic context established in the text, knowledge of the causal relationships for the domain in question, and the communicative goals of the text’s author (e.g., recognition of a well known causal relationship or conveying a possibly new causal relationship)

Causality in the text is expressed in arbitrarily complex ways; even when events are in a single sentence, it might be mentioned in a sparse, ambiguous, or implicit manner. Even explicit causal expressions have rich variation across their syntax, and it is challenging to extract complete lexical features and understand causality based on any single grammatical model. Figure~\ref{fig:causality-in-text} shows a few popular classes of explicit causality as discussed by \textit{Khoo et al.}\cite{Khoo2000}. 
For example, in the sentence "smoking causes cancer," it is easy to understand the causal relationship between smoking and cancer. Nevertheless, it gets complicated when the interaction between events is implicit. In the sentence, "I think the greater danger is from the unelected justices than from the elected Congress and the elected president." Identifying the "Cause-Effect" interaction between "greatest danger" and "unelected justices" from the given text is not a trivial task. The task gets complicated when the "Cause-Effect" is not explicitly mentioned between events of interest. Furthermore, as the length of the sentence increases or there is more than one pair of casual events in a given sentence, it becomes difficult for any statistical model to capture the sentence's context unambiguously. 
Causality in the text can be expressed simply by using the prepositions of the form "A causes B" or "A is caused by B." This form of expression is intuitive and can often be subjective to the field and also to the researcher's views.
Instead of predicting "A causes B," in this work, we predict "could A cause B" from a given textual expression, where causal interaction between events can be explicit or implicit with
a wide variety of writing styles. 

\section{Dataset}\label{dataset}
We used three different datasets to train and evaluate our models; Semeval 2007 and Semeval 2010 is curated using pattern-based web search while ADE is curated from biomedical text. Semeval 2007, Semeval 2010, and  ADE datasets are publicly available and have annotated events and "Cause-Effect" interaction between them. 
Unlike \textit{Dunietz et al. and Mirza et al.} \cite{dunietz2017because}, \cite{mirza-etal-2014-annotating}, who 
incorporated the association between temporal and other relations with causality
and  \textit{Prasad et al} \cite{prasad2008penn}, who focuses on inter-sentence causality; in this task, we only focused on causality between events in the same sentence. 

Table~\ref{table:data-details} provides statistical details about datasets and Table~\ref{table: data-description} provides a sample example from each of the curated dataset. Further details on data curation and preparation are below:  


\noindent - \textbf{SemEval2007:} SemEval 2007 \cite{Girju2007} is an evaluation task designed to provide a framework for comparing different approaches to classifying semantic relations between nominals in a sentence. The task provides a dataset for seven different semantic relations, including "Cause-Effect". For this work, we use part of the SemEval 2007 dataset with the "Cause-Effect" relationship. For a given sentence, if the interaction between marked events is causal, we label it as "Cause-Effect" else the sentence is labeled as "Other." \\
\noindent - \textbf{SemEval2010:} Similar to the above dataset, we use SemEval 2010  \cite{Hendrickx2010} dataset with causal interaction between events labeled as "Cause-Effect", and all the other types of interactions between events in rest of the sentences are labeled as "Other". \\
- \textbf{ADE:} The adverse drug effect \cite{Gurulingappa2012} dataset is a collection of biomedical text annotated with drugs and their adverse effects. The first corpus of this dataset, with drugs causing adverse effects, has drugs as well as effects annotated. In the second corpus, where drugs are not causing any side-effect, the drug and its effect name are not manually annotated. We curated a list of unique drugs and affect names using the first corpus data and use this set to annotate the drugs and effect names in the second corpus.  While we take sentences with two or more drugs/effect mention in them; for simplicity, we do not replicate the sentence in our final corpus. We marked the first two mention of drug/effect mention we find in the sentence.\\


\begin{table}[!htbp]
\centering
\caption{Statistics for curated datasets}
\begin{tabular}{|l|l|l|l|l|l|l|l|} 
\hline
             &                                                                            & \multicolumn{3}{l|}{\textbf{Train Dataset}} & \multicolumn{3}{l|}{\textbf{Test Datast}}    \\ 
\hline
Dataset      & \begin{tabular}[c]{@{}l@{}}Max sentence length\\(train, test)\end{tabular} & \#Total & \#Cause-Effect & \#Other & \#Total & \#Cause-Effect & \#Other  \\ 
\hline
Semeval 2010 & (85, 60)                                                                   & 8000    & 1003           & 6997    & 2717    & 134            & 2389     \\ 
\hline
Semeval 2007 & (82,62)                                                                    & 980     & 80             & 900     & 549     & 46             & 503      \\ 
\hline
ADE          & (135, 93)                                                                  & 8947    & 5379           & 3568    & 2276    & 1341           & 935      \\
\hline
\end{tabular}
\label{table:data-details}
\end{table}

\begin{table*}[!htbp]
 \caption{Example sentences, sentence with event markers, and masked event markers for curated datasets}

\begin{tabular}{ |p{1.5cm}|p{4cm}|p{4cm}|p{4cm}|  }
 \hline
 \multicolumn{4}{|c|}{\textbf{Curated Corpus}} \\
 \hline
 Dataset & Example Sentence 	&Sentence with event marker & Sentence with masked event marker\\
 \hline
 Semeval 2007 & Most of the taste of strong onions comes from the smell.	&Most of the \textbf{$<$e1$>$ taste $<$/e1$>$} of strong onions comes from the \textbf{$<$e2$>$ smell$<$/e2$>$}.&Most of the \textbf{$<$e1$>$ blank $<$/e1$>$} of strong onions comes from the \textbf{$<$e2$>$ blank $<$/e2$>$}.\\
  \hline
 Semeval 2010 &	As in the popular movie "Deep Impact", the action of the Perseid meteor shower is caused by a comet, in this case periodic comet Swift-Tuttle. & As in the popular movie "Deep Impact", the action of the Perseid \textbf{$<$e1$>$ meteor shower $<$/e1$>$} is caused by a \textbf{$<$e2$>$ comet $<$/e2$>$}, in this case periodic comet Swift-Tuttle.&  As in the popular movie "Deep Impact", the action of the Perseid \textbf{$<$e1$>$ blank $<$/e1$>$} is caused by a \textbf{$<$e2$>$ blank $<$/e2$>$}, in this case periodic comet Swift-Tuttle.\\
  \hline
 ADE	& Quinine induced coagulopathy --a near fatal experience.	& \textbf{$<$e2$>$ Quinine $<$/e2$>$} induced \textbf{$<$e1$>$ coagulopathy $<$/e1$>$}--a near fatal experience.&	\textbf{$<$e2$>$ blank $<$/e2$>$} induced \textbf{$<$e1$>$ blank $<$/e1$>$}--a near fatal experience.\\
  \hline
 \end{tabular}

 \label{table: data-description}
 \end{table*}


\begin{table}[!htbp]
\caption{Experiment Setup}
\centering
\begin{tabular}{|l|l|l|l|l|l|} 
\hline
Optimizer & Learning rate & Epsilon & Dropout rate & Train batch & Max sequence length  \\ 
\hline
Adam      & 1e-05         & 1e-08   & 0.4          & 16          & 384                  \\
\hline
\end{tabular}
\label{expr table}
\end{table}
\section{Problem definition and Proposed Methodology}
Our causality understanding approach can be simplified as a binary classification of "Cause-Effect"/"Other" relationship between events expressed in natural language text. The detailed methodology, problem definition and network architecture are described below: 
\subsection{Methodology}
Our methodology involves:
\begin{itemize}
\item Fine-tuning Bert based feed forward network for ``Cause-Effect"/``Other" relationship label between events expressed in natural language text
\item Combining both the event's context and BERT's sentence context to predict ``Cause-Effect"/``Other" relationship label between events. This methodology is built on the method suggested by \textit{Wu et al.}\cite{Wu2019} and works on the intuition that the interaction between two events is result of the information in the sentence as well as in the events
\item Combining both the event's masked context with BERT's sentence context to predict ``Cause-Effect"/``Other" relationship label between events. This methodology is built on the model suggested by \textit{Soares et al.}\cite{Soares2020} and works on the intuition that training with event's masked context can help language model learn task specific implicit sentence structure(\cite{Jawahar2020} and \cite{Gururangan2020}). 

\end{itemize}
\subsection{Problem definition}
The overall problem can be defined as follow: for a given sentence and the two marked events $\mathbf{e_1, e_2}$, the goal is to predict the possible casual interaction $\mathbf{c}$ between the events. Mathematically, a given sentence can be seen as a sequence of tokens $\mathbf{x}$, and possible interaction $\mathbf{c}$ can be either Cause-Effect or Other. Event$_1$ and Event$_2$ are a continuous span of text in the given statement.

For a given sequence of token 
\begin{equation}
    \mathbf{x = [x_0,x_1,x_2,\cdots,x_n]}
\end{equation} 
\begin{equation}
    \mathbf{event_1  : E_1 = [x_i, x_j]}
\end{equation}
\begin{equation}
    \mathbf{event_2  :  E_2 = [x_k, x_l]}
\end{equation}
\begin{equation}
    \mathbf{relation   :  c\in [Cause-Effect, Other]}
\end{equation}
0$<$i$<$j-1;  j$<$k; k$\leq$l-1; l$\leq$n, where n is sequence length.\\

For a given sentence \textbf{x} with two marked Event$_1$ and Event$_2$, we used a pre-trained BERT model to obtain the event expression encoding h$_r$ = R$^d$ in the contextualized latent space. This latent space representation varies based on the model used for experimentation. Finally, the model is trained to maximize the probability (P(interaction$|$h$_r$)) of causal interaction between event for a given context vector (h$_r$ = R$^d$). During Inference, the trained model predicts the possible interaction based on marked events and sentence expression.
\subsection{Models}
\label{Proposed Methodology:Models}
BERT \cite{Devlin2019BERT:Understanding} is a pre-trained language model based on a widely used Transformer architecture proposed by \textit{Vaswani et al.}\cite{Vaswani2017}. BERT provides deep bidirectional representations from the unlabeled text by jointly conditioning on the left and right contexts. BERT based pre-trained models has shown to be very effective at many tasks, including question-answering, relation extraction, and NLI. We built the following three network architectures on top of the  pre-trained BERT\footnote{BERT diagrams - adapted from the image by Jimmy Lin} model to investigate and evaluate our proposed methodologies:\\

\begin{figure}
\centering
\begin{minipage}{.5\textwidth}
  \centering
  \includegraphics[width=\linewidth,keepaspectratio]{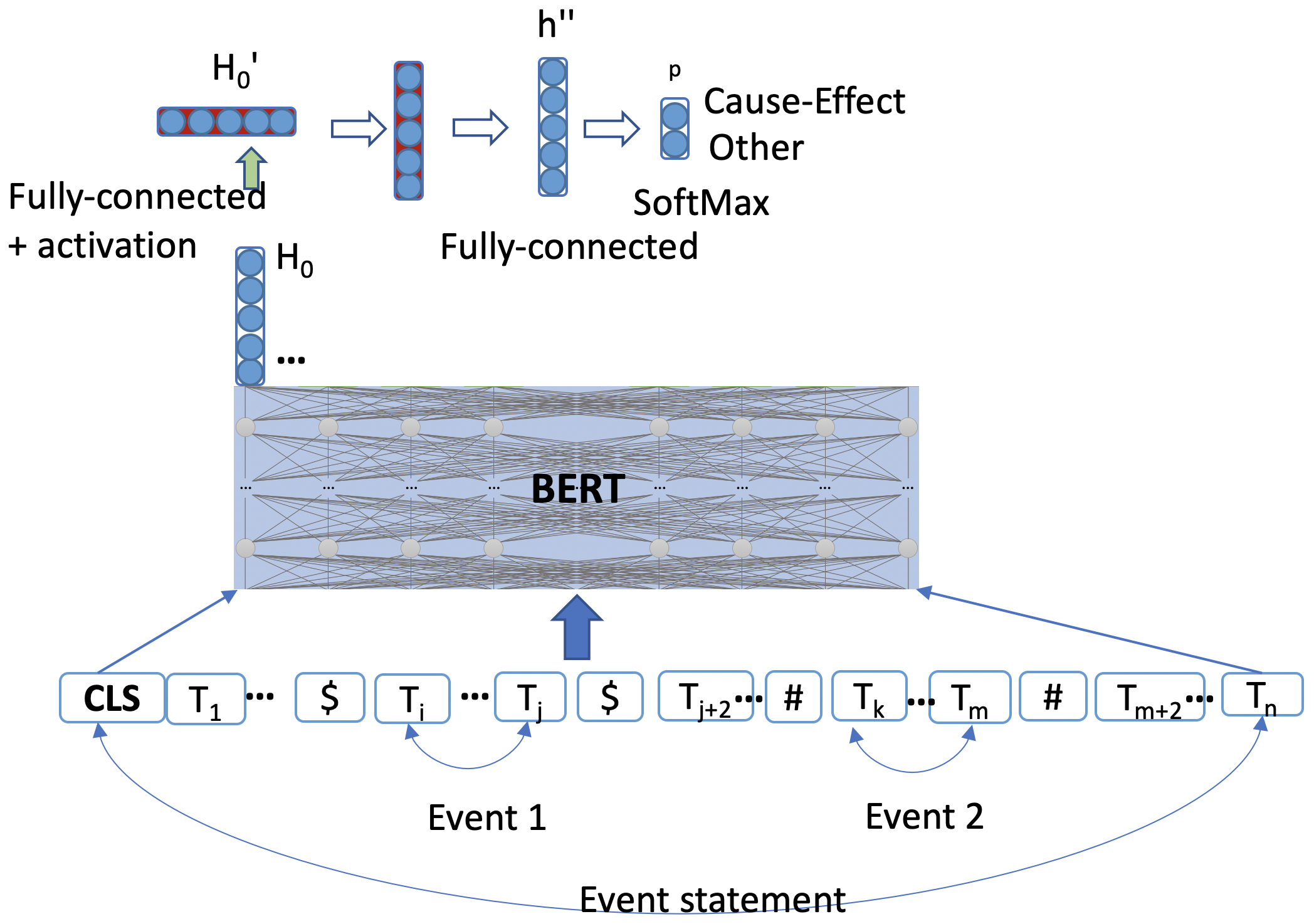}
  \caption{C-BERT}
  \label{fig:cbert}
\end{minipage}%
\begin{minipage}{.5\textwidth}
  \centering
  \includegraphics[width=\linewidth,keepaspectratio]{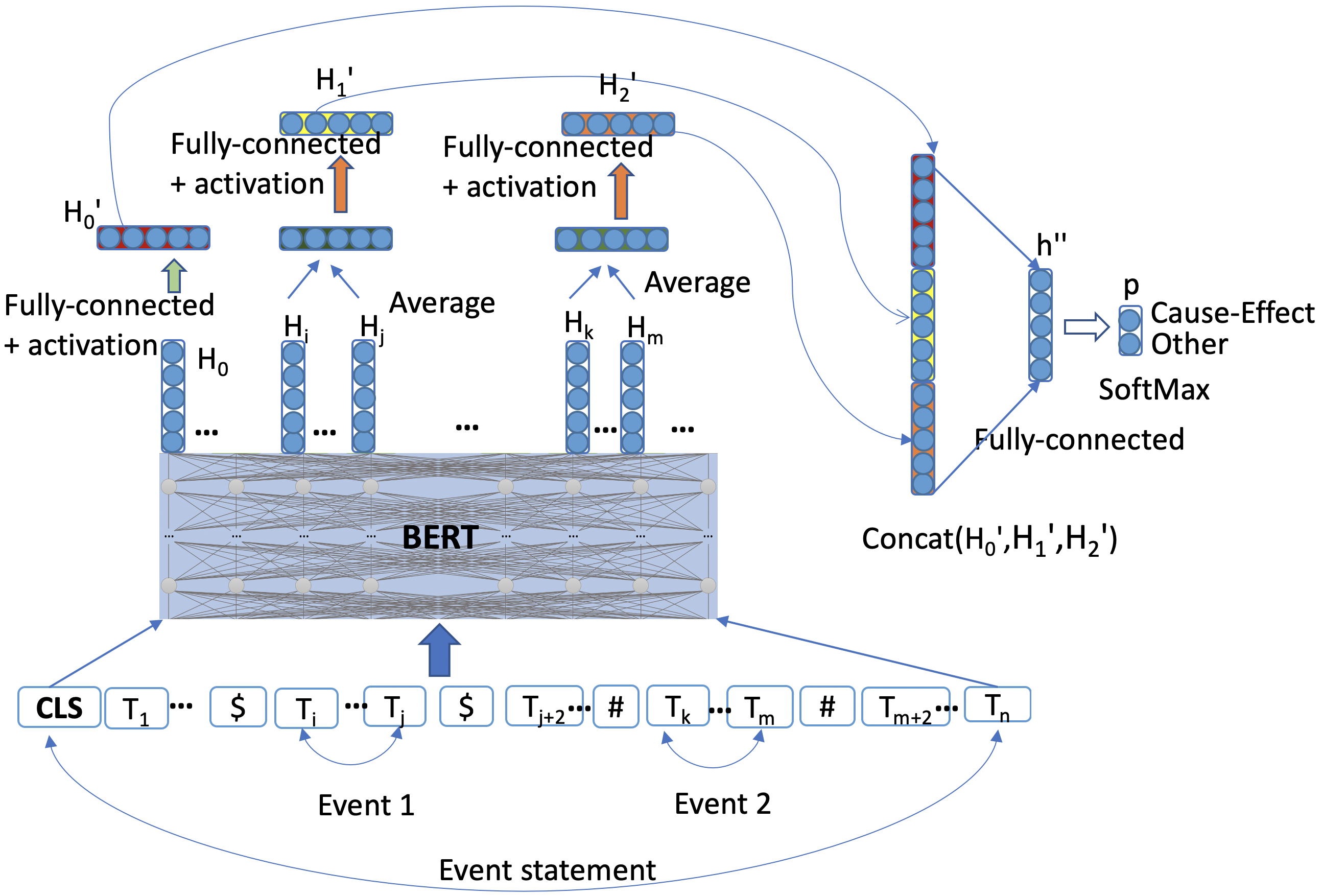}
  \caption{Event aware C-BERT}
  \label{fig:event aware cbert}
\end{minipage}
\end{figure}

\begin{itemize}
    \item \textbf{C-BERT:} \label{C-BERT}
    This network architecture is a feed-forward network build on top of BERT \cite{Devlin2019BERT:Understanding}. This network can be trained for binary classification of the ``Cause-Effect"/``Other" relationship between two events in a given input sentence. In this network architecture, we feed the input sentence as a sequence of tokens to the BERT model and take the overall sentence context vector from the BERT model's output, feed it to a non-linear activation layer followed by two fully connected layers. We have applied dropout before each fully connected layer with a probability of 0.4. The likelihood of interaction being Cause-Effect using a soft-max layer is shown in fig:~\ref{fig:cbert}. We used back-propagation with adam-optimizer on binary loss function to learn the optimal solution with training batch size as 16. 
    \noindent Mathematical formulation for C-BERT model is:
    \begin{equation}
    H_0' = W_0(tanh(H_0)) + b_0
    \end{equation}
    \begin{equation}
    h'' = W_1(H_0')+b_1    
    \end{equation} 
    \begin{equation}
    p = softmax(h'')
    \end{equation}
    where, $W_0\in R^{d\times d}$; $W_1 \in R^{L\times d}$, $H_0$ is the output token of bi-directional context (i.e. [CLS]) of BERT, and L = 2 (Cause-Effect, Other)\\
        
    \item \textbf{Event aware C-BERT:}\label{Event aware C-BERT}
    
    This network architecture learns event informed representation of the given sentence expression and can predict causality between identified/marked events.  As shown in the fig:~\ref{fig:event aware cbert}, events can be more than a single token, resulting in many vectors when the input sentence is fed into a pre-trained BERT model. We averaged them to get the final context of each event expression and passed the sentence context as well as both the event's context to a non-linear activation layer followed by a fully connected layer. The sentence context is concatenated with both the events' averaged context and is feed to another fully connected layer followed by a softmax layer. This network architecture also has a dropout before each fully connected layer with a probability of 0.4. Finally, the model is trained using back-propagation with adam-optimizer on a binary loss function to predict the ``Cause-Effect"/"Other" relationship between events. 
    \noindent Mathematical formulation for Event Aware C-BERT model is : 
    \begin{equation}
    H_0' = W_0(tanh(H_0)) + b_0    
    \end{equation}
    \begin{equation}
    H_1' = W_1\left[ \frac{1}{j-i+1}\sum_{t=i}^{j}tanh\left(H_{t} \right)\right ]+b_1   
    \end{equation}
    \begin{equation}
    H_2' = W_2\left[ \frac{1}{j-i+1}\sum_{t=i}^{j}tanh\left(H_{t} \right)\right ]+b_2   
    \end{equation}
    \begin{equation}
    h'' = W_3\left[concat(H_0',H_1',H_2') \right ]+b_3    
    \end{equation} 
    \begin{equation}
    p = softmax(h'')    
    \end{equation}
    where, $W_0$, $W_1$, $W_2 \in R^{d\times d}$; $W_3 \in R^{L\times 3d}$ and L
    = 2 (Cause-Effect, Other).\\
    \item \textbf{Masked Event C-BERT:}\label{Masked Event C-BERT}
    
    This network architecture shown in fig:~\ref{fig:masked event cbert} is very similar to the event aware C-BERT network architecture (fig:~\ref{fig:event aware cbert}), where the whole span of event text is replaced with a "BLANK" token. As each event is just a single blank token, unlike Event aware C-BERT we don't need to take an average to get the final context of any event. Each model trained by this approach is then fine-tuned using actual event information using the Event Aware C-BERT model described above.
    \noindent Mathematical formulation for Masked Event C-BERT is:
    \begin{equation}
    H_0' = W_0(tanh(H_0)) + b_0    
    \end{equation}
    \begin{equation}
    H_1' = W_1\left[ tanh\left(H_{ij} \right)\right ]+b_1   
    \end{equation}
    \begin{equation}
    H_2' = W_2\left[ tanh\left(H_{km} \right )\right ]+b_2    
    \end{equation}
    \begin{equation}
    h'' = W_3\left[ concat(H_0',H_1',H_2') \right ]+b_3    
    \end{equation}
    \begin{equation}
    p = softmax(h'')    
    \end{equation}
    where, $W_0$, $W_1$, $W_2 \in R^{d\times d}$; $W_3 \in R^{L\times 3d}$ and L 
    = 2 (Cause-Effect, Other).
    
\end{itemize}

\section{Experiments and results}
To evaluate the language model's capabilities in learning the ``Cause-Effect" relationship between events, we trained three different BERT \cite{Devlin2019BERT:Understanding} based network architecture's on each one of the dataset's. We also experimented with the effectiveness of pretraining a Masked Event C-Bert model on the out of domain data distribution and fine-tuning using the Event Aware C-BERT model for handling low resource situations. For each of the dataset distribution, we created two different corpora. The first corpus is created by adding a positional indicator before, and after both the event in the sentence, and then to prepare the second corpus, we replaced the event text span with a BLANK token. 
\subsection{Experiments:}
\begin{figure}
\centering
\begin{minipage}{.46\textwidth}
  \centering
  \includegraphics[width=\linewidth,keepaspectratio]{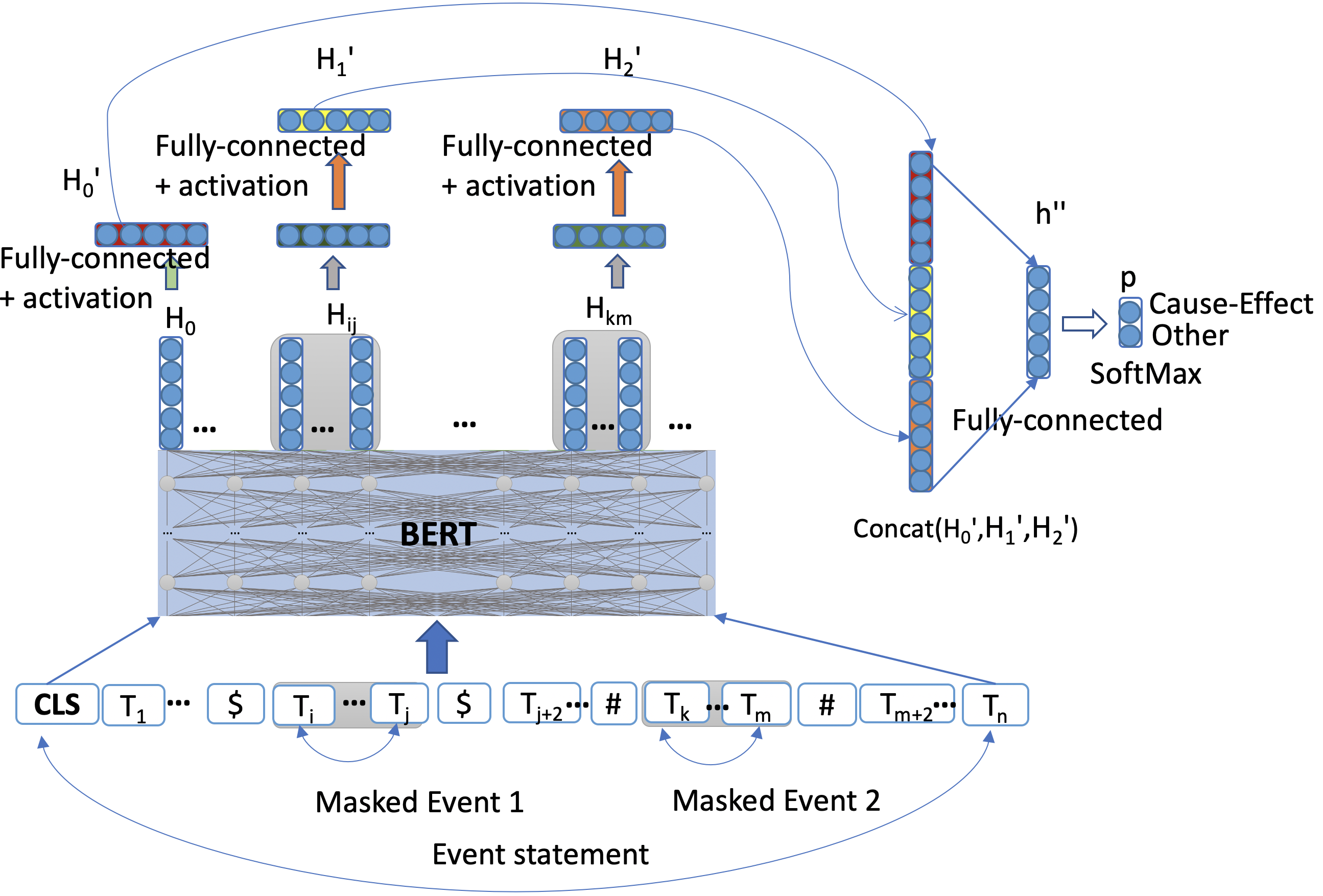}
  \caption{Masked Event C-BERT}
  \label{fig:masked event cbert}
\end{minipage}%
\begin{minipage}{.54\textwidth}
  \centering
  \includegraphics[width=\linewidth,keepaspectratio]{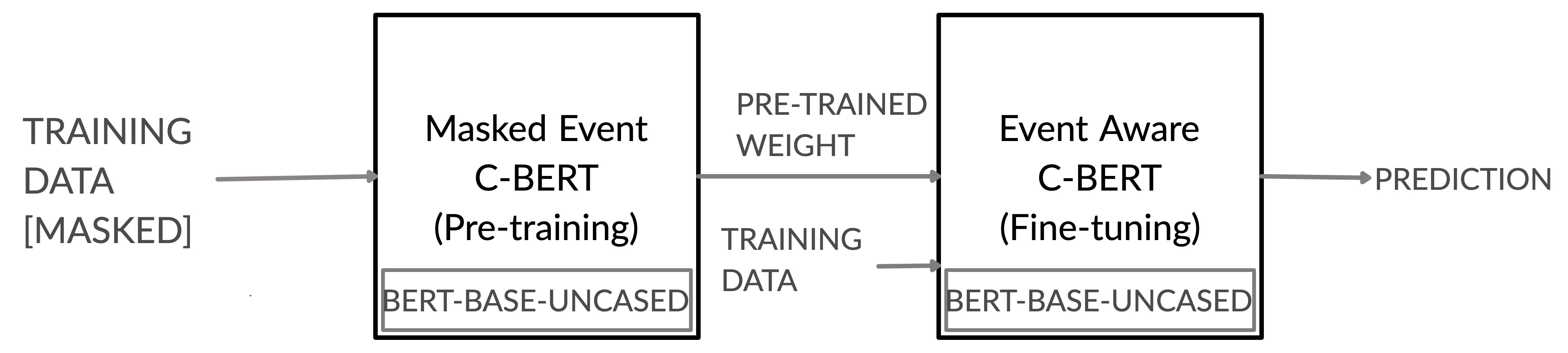}
  \caption{Pre-training on Masked Event C-BERT and Fine-Tuning on Event Aware C-BERT}
  \label{fig:pretrain-finetuning}
\end{minipage}
\end{figure}

Using models described in Section Models, we performed four sets of experiments for each of the dataset. The experiments are discussed in more details below: 

\begin{itemize}
    \item End to end training of C-BERT model using target data distribution: In these set of experiments, for each of the data distribution, we trained a model based on C-BERT's network architecture. For each experiment, the input is a sequence of tokens, and the output is the "Cause-Effect"/"Other" relationship between them. This is our base model and reported performance gain on the previously reported results by \textit{Girju et al. and Dasgupta et al.}[\cite{Girju2002}, \cite{Dasgupta2019}].
    \item End to end training of Event Aware C-BERT model using target data distribution: In these set of experiments, for each of the data distribution, we trained a model based on Event Aware C-BERT's network architecture. In each of these training, the network combines the BERT's bidirectional context of the sentence with both the event's encoding. These experiments improved our base model's performance for all three data distribution.
    \item Pre-training of Masked Event C-BERT and fine-tuning of Event aware C-BERT both using same in-domain data distribution: In these set of experiments, for each of the data distribution, we pre-trained a model based on Masked Event C-BERT's network architecture and fine-tuned the weights using Event Aware C-BERT's network architecture. Here, by pre-training on masked event corpus, we leverage the language model's ability to learn implicit task-specific sentence argument representation as discussed by \textit{Jawahar et al. and Gururangan et al.}[\cite{Jawahar2020},\cite{Gururangan2020}]. When trained (pre-training + fine-tuning) the models using only in-domain data distribution and obtained either similar or improved performance gain on all four data distribution.
    \item Pre-training of Masked Event C-BERT using a different data distribution than our target data distribution and fine-tuning of Event Aware C-BERT using target data distribution: Task of ``Cause-Effect" prediction between events does not have a large scale dataset and is as challenging as all other low resource language tasks. Moreover, it is also challenging to transfer trained models from one data distribution to another (even in the same domain) or from one domain to another. Based on the findings of \textit{Jawahar et al.}\cite{ Jawahar2020}, in these experiments, we tried to learn implicit task-specific sentence argument representation  by pre-training on distribution different from our target data distribution. For each of the available data distribution, we pre-trained two models using the other two out of domain distribution, respectively.  This set of experiments resulted in additional performance gain for most of the experiments and is shown in table \ref{evaluation table}. Fig: \ref{fig:pretrain-finetuning} gives an intuitive overview of the proposed training methodology.

\end{itemize}







\subsection{Learning setup: experiment settings }
We used AWS p3.2xlarge instance for all the experiments. It has 8 vCPU and 61 GiB Memory and NVIDIA Tesla V100 GPUs with 16 GiB Memory. To test the performance gain with the change in network architecture, we used the same hyper-parameter for all the experiments. 

\subsection{Results and analysis}
In this work, we experimented with network architecture built on top of BERT to investigate language models' capability in causality understanding between events expressed in natural language text. 

Table~\ref{Table:PRF1} compares the performance of our models built using three different network architecture trained on in/out of domain data distribution with previously reported F1 performance measures. 

Our base C-BERT model itself outperformed the previously reported performance measure \cite{Dasgupta2019, Girju2007, Huynh2016AdverseDR} on the corpus created using the SemEval 2010, SemEval 2007, and ADE dataset. Further experimentation using event aware C-BERT and training methodology described in fig~\ref{fig:pretrain-finetuning} (Masked Event C-BERT for pre-training and Event Aware C-BERT for fine-tuning on the same dataset) reported additional performance gain, respectively.

Table~\ref{evaluation table} shows the result of another set of experiments where we evaluate the performance of the models when pre-training and fine-tuning is performed using in-domain data distribution in comparison to the models when pre-training uses out of domain data distribution. For each of the target data distribution, we pre-trained three models using the other out of domain data distribution. In general, pre-training on a different dataset than our target data distribution reported either similar or improved performance.

Further analysis of the results shows that Semeval 2007 benefited most from our experimental settings. Semeval 2007 is the smallest of the three corpora. It gained more performance than others when models were pre-trained on out of domain data-distribution. Therefore, one can intuitively understand that pre-training helps the model learn implicit task-specific sentence's lexical representation in line with the findings of \textit{Jawahar et al. and Gururangan et al.} \cite{Jawahar2020, Gururangan2020}.


\begin{table}
\centering
\caption{Comparison of F1 Score of models on SemEval 2007, SemEval 2010, Adverse Drug Effect. All experiments are performed using only in domain data distribution. Entries in column Girju et al. (2007) is weighted over all 7 classes as reported by them.}
\begin{tabular}{|l|l|l|l|}
\hline
                                                                                   & Semeval 2007           & Semeval 2010 & ADE   \\ \hline
C-BERT                                                                             & 93.78                  & 97.68        & 97.10 \\ \hline
Event Aware C-ERT                                                                  & 94.94                  & 98.35        & 97.85 \\ \hline
\begin{tabular}[c]{@{}l@{}}Masked Event C-BERT +\\ Event Aware C-BERT\end{tabular} & 95.31                  & 97.85        & 97.85 \\ \hline
Dasgupta et al.(2019)                                                              & \multicolumn{1}{c|}{-} & 75.39        & -     \\ \hline
Girju et al.(2007)                                                                 & 82.00                  & -            & -     \\ \hline
Huynh et al.(2016)                                                        & -                      & -            & 87.00 \\ \hline
\end{tabular}
\label{Table:PRF1}
\end{table}

\begin{table}
\centering
\caption{F1 Score after pre-training on masked event C-BERT model and fine-tuning on Event Aware C-BERT}

\begin{tabular}{|l|l|l|l|l|} 
\hline
                                                                                                 & \multicolumn{4}{l|}{Dataset for fine-tuning of event aware C-BERT model}  \\ 
\hline
\begin{tabular}[c]{@{}l@{}}Dataset for pre-training of \\masked event C-BERT model \end{tabular} &              & Semeval 2007 & Semeval 2010 & ADE                          \\ 
\hline
{}                                                                                & Semeval 2007 & 95.31        & 98.42        & 97.27                        \\ 
\cline{2-5}
                                                                                                 & Semeval 2010 & 97.14        & 98.39        & 97.47                        \\ 
\cline{2-5}
                                                                                                 & ADE          & 96.42        & 98.49        & 97.85                        \\
\hline
\end{tabular}
\label{evaluation table}
\end{table}
\section{Related work}
One direction of efforts in causality detection between events expressed in natural language text can be summarised as the use of lexico-syntactic patterns and contextual cue matching around the events. Early proposed methods in this direction lacked coverage and were domain-dependent \cite{Gaxcia1997}. \textit{Khoo et al.}\cite{Khoo2000} proposed a syntactic graph pattern matching method for explicitly mentioned "Cause-Effect" relationships between events and reduced a lot of domain knowledge requirements. An interesting semi-automatic linguistic pattern-based approach came from \textit{Girju et al.}\cite{Girju2002}. They automatically extract causal relationships based on explicit inter-sentence lexico-syntactic pattern of the form $<NP1$ verb $NP2>$ and then disambiguate the extracted causal relationship based on semantic constraints on nouns and verbs. Later \textit{Girju et al.} \cite{Girju2003} also showed the usefulness of the causation module for QA tasks, especially when the questions are ambiguous and implicit. In contrast to the above work, we are trying to learn causality without doing extensive feature engineering. 

Another direction of efforts can be summarised as data-driven approaches. Early proposed methods in this direction use Point-wise Mutual Information (PMI) \cite{Church1990}. \textit{Mihalcea et al.}\cite{Mihalcea2006} proposed a PMI based semantic similarity method that measures semantic similarity based on the co-occurrence of frequent pairs of words in the given sentence. But, this method results in noisy predictions depending upon the variation in the frequency of the words. The Cause-Effect association (CAE) method proposed by \textit{Do et al.}\cite{Do2011} improves their similarity measure by penalizing phrases that occur frequently. However, this method ignored the semantics of text spans surrounding the causal events. \textit{Hu et al.}\cite{Hu2017} proposed a Causal Potential (CP) based method to access causal relation between everyday events (film-scripts, blogs) considering the relative ordering of the event. 

An interesting data-driven approach using support vector machines(SVMs) was suggested by \textit{Beamer et al.}\cite{Beamer2007}. They used a knowledge-driven approach to extract lexical, syntactic, and semantic features for automatic identification of semantic relations in SemEval 2007 Task 04 \cite{Girju2007}. Another interesting data-driven approach was suggested by \textit{Dasgupta et al.}\cite{Dasgupta2019}. They leveraged a bi-directional LSTM model for general-purpose causality detection. However, it lacks the generality of language models and will not give expected results when a sequence of unseen natural language expression is fed. \textit{Dasgupta et al.}\cite{Dasgupta2019} also proposed the use of a lexical feature-based k-means clustering to cluster Cause-Effect events. However, \textit{Sharp et al.}\cite{Sharp2016} had used the method suggested by \textit{Goldberg et al.}\cite{Goldberg2019} to replace the standard linear context with the causal dependency context, resulting in a more task-specific and context dedicated embedding for Cause-Effect event detection. These proposed data-driven approaches either don't use the linguistic context of the event expression or lacks the complete bi-directional understanding of the text expression. Our proposed approach uses BERT's bidirectional context to learn causal interaction between events.  

More recently,\textit{Hassanzadeh et al.}\cite{Hassanzadeh2019} used BERT based language model to identify and rank sentences similar to binary casual questions ("If A could cause B."). They used ranked documents to add explainability to their methods' final prediction and also to get over the restriction on what type of phrases they can use or the need to map the causal phrases to some knowledge base. \textit{Hassanzadeh et al.} \cite{Hassanzadeh2020} also showed causal embedding and language modeling-based approaches could be useful in extracting causal knowledge from a large scale unstructured text. We also used BERT based language models for our causality understanding task but unlike \textit{Hassanzadeh et al.}, in this work, we predicted implicit and explicit causal relationship between previously identified events in natural language text.

\section{Conclusion}
In this work, we studied the problem of causal relationship detection between events expressed in natural language text where both the events are expressed as nominal or phrase or a short span of text in the same sentence. We showed that the network architectures built on top of the contextualized language model can learn causal relations in the text using sentence context, event information, and masked event context. Our proposed methodology significantly outperforms the reported performance measure on curated datasets, and it can be useful in automatic causal graph extraction as well as identifying a causal chain of events in event expressions. This can further help recognize counterfactuals by connecting antecedent and consequent with causal relations. 

For a complete causal understanding of events expressed in natural language text, we need the ability to recognize sentences with causal events, identify those events, causal relations between the events, and understand the influences between those events. There can be several future directions from here. One interesting direction would be the automatic identification of events as well as sub-events in a given text expression. And then, identifying richer and fine-grained causal relationships (e.g., caused, contributed, likelihood, and encouraged, influenced) between them even when the events are described in the same or different sentences.

\bibliographystyle{plain}
\bibliography{causal-BERT.bib}
\newpage
\appendix
\section{Appendix}


In this section, we elaborate on data curation with examples and showed few example where our models worked as well as where it missed totally. 

Table ~\ref{Table:what-worked-and-dont} shows examples where our model could detect causality in sentences where it was not apparent as well as where our model got confused. Table ~\ref{Table:what-worked-and-dont} shows two examples of causal sentences where event interactions were correctly classified as Cause-Effect (True Positive) and as Other (True Negative) and where event interactions were incorrectly classified as Cause-Effect (False Positive) as Other (False Negatives).

\begin{table}[!htbp]
\centering
\caption{Example sentences for True Positive/Negative and False Positive/Negative}
\begin{tabular}{|p{2cm}|p{10cm}|}
\hline
{\begin{tabular}[c]{@{}l@{}}True \\ positives\end{tabular}}  & This as well as kinetic data support the hypothesis of \textless e1 \textgreater inhibition \textless e1\textgreater through \textless e2 \textgreater altered membrane properties \textless /e2 \textgreater.                                              \\ \cline{2-2} 
                                                                            & \textless{}e1\textgreater Dehydration \textless{}/e1\textgreater from \textless{}e2\textgreater fluid loss \textless{}/e2\textgreater generally is the only major problem the virus can cause, most often in the elderly.                                       \\ \hline
{\begin{tabular}[c]{@{}l@{}}False \\ Positives\end{tabular}} & If your main \textless{}e1\textgreater aim \textless{}/e1\textgreater is for \textless{}e2\textgreater growth \textless{}/e2\textgreater in the value of the property then obviously you need to look at where you think the next "value-spurt" is going to be. \\ \cline{2-2} 
                                                                            & \textless{}e1\textgreater Brushing \textless{}/e1\textgreater after \textless{}e2\textgreater meals \textless{}/e2\textgreater should become part of your daily schedule.                                                                                       \\ \hline
{\begin{tabular}[c]{@{}l@{}}True \\ Negatives\end{tabular}}  & My \textless{}e1\textgreater sore throat \textless{}/e1\textgreater from \textless{}e2\textgreater yesterday \textless{}/e2\textgreater has turned into a full-blown cold overnight.                                                                            \\ \cline{2-2} 
                                                                            & The \textless{}e1\textgreater evacuation \textless{}/e1\textgreater after the Chernobyl \textless{}e2\textgreater accident \textless{}/e2\textgreater was poorly planned and chaotic.                                                                           \\ \hline
{\begin{tabular}[c]{@{}l@{}}False \\ Negatives\end{tabular}} & Most of the \textless{}e1\textgreater taste \textless{}/e1\textgreater of strong onions comes from the \textless{}e2\textgreater smell \textless{}/e2\textgreater{}.                                                                                            \\ \cline{2-2} 
                                                                            & Subjects will be secured at all times with a safety harness to prevent \textless{}e1\textgreater injury \textless{}/e1\textgreater from \textless{}e2\textgreater falling \textless{}/e2\textgreater{}.                                                         \\ \hline
\end{tabular}
\label{Table:what-worked-and-dont}
\end{table}
Table ~\ref{Table:total_unique_sentence} shows the example sentence as well as marked events in them. In Semeval 2007 and Semeval 2010 datasets, original data contained additional information about the direction of causality. But due to the small size of the dataset our experiments do not focused on the directional aspects of causal interaction between events. In Semeval 2007 and Semeval 2010 dataset, we only had one pair of events in each sentence, whereas, in ADE, we had more than one pair of events in each sentence. 

While curating the data for our experiments, for each sentence, for each pair of interactions, we curated a new data point. In the Semeval 2007 and Semeval 2010 ratio of unique lines to total lines is 1, whereas, in the ADE dataset it is 0.978$/$0.975 (test/train). It is clear that ADI has some repeating sentence expressions with different pairs of marked events.
\begin{table}[!htbp]
\centering
\caption{Example of a sentences with event pair for each dataset}
\begin{tabular}{|p{2.0cm}|p{6.0cm}|p{3.5cm}|}
\hline
\textbf{Dataset} & \textbf{Sentences} & \textbf{Event Pairs} \\ \hline
Semeval 2007 & He had chest pains and \textbf{headaches} from \textbf{mold} in the bedrooms. & (headaches, mold) \\ \hline
Semeval 2010 & The \textbf{treaty} establishes a double majority \textbf{rule} for Council decisions. & (treaty, rule) \\ \hline
ADE & A 79-year-old man with ischemic heart disease, chronic atrial fibrillation, chronic renal failure, hypothyroidism, and gout arthritis was hospitalized because of \textbf{fatigue}, \textbf{myalgia}, and \textbf{leg weakness}, shortly after starting treatment with \textbf{colchicine}. & (fatigue, colchicine) (myalgia,colchicine) (leg weakness, colchicine) \\ \hline

\end{tabular}

\label{Table:total_unique_sentence}
\end{table}

\end{document}